\documentclass{article}

\usepackage{PRIMEarxiv}

\usepackage[utf8]{inputenc} 
\usepackage[T1]{fontenc}    
\usepackage{hyperref}       
\usepackage{url}            
\usepackage{booktabs}       
\usepackage{amsfonts}       
\usepackage{nicefrac}       
\usepackage{microtype}      
\usepackage{lipsum}
\usepackage{fancyhdr}       
\usepackage{graphicx}       
\graphicspath{{media/}}     

\usepackage{algorithmic,algorithm,xpatch}

\xpatchcmd{\algorithmic}{\setcounter}{\algorithmicfont\setcounter}{}{}
\providecommand{\algorithmicfont}{}

\usepackage{amsmath} 
\usepackage{subcaption}
\usepackage{enumitem}
\usepackage{array}
\usepackage{tabularx}
\usepackage{multirow}

\title{Low-Resolution Neural Networks
}

\author{
  Eduardo~Lobo~Lustosa~Cabral \\
  IPEN - Institute for Energy and Nuclear Research \\
  Mauá Institute of Technology, Engineering School\\
  São Paulo-SP, Brazil\\
  \texttt{elcabral@ipen.br,elcabral@maua.br} \\
   \And
  Larissa~Driemeier \\
    University of São Paulo \\
    Polytechnic School, Department of Mechatronics and Mechanical Systems Engineering\\
  São Paulo-SP, Brazil\\
  \texttt{driemeie@usp.br} \\
}

\begin{document}
\maketitle

\begin{abstract}
The expanding scale of large neural network models introduces significant challenges, driving efforts to reduce memory usage and enhance computational efficiency. Such measures are crucial to ensure the practical implementation and effective application of these sophisticated models across a wide array of use cases. This study examines the impact of parameter bit precision on model performance compared to standard 32-bit models, with a focus on multiclass object classification in images. The models analyzed include those with fully connected layers, convolutional layers, and transformer blocks, with model weight resolution ranging from 1 bit to 4.08 bits. The findings indicate that models with lower parameter bit precision achieve results comparable to 32-bit models, showing promise for use in memory-constrained devices. While low-resolution models with a small number of parameters require more training epochs to achieve accuracy comparable to 32-bit models, those with a large number of parameters achieve similar performance within the same number of epochs. Additionally, data augmentation can destabilize training in low-resolution models, but including zero as a potential value in the weight parameters helps maintain stability and prevents performance degradation. Overall, 2.32-bit weights offer the optimal balance of memory reduction, performance, and efficiency. However, further research should explore other dataset types and more complex and larger models. These findings suggest a potential new era for optimized neural network models with reduced memory requirements and improved computational efficiency, though advancements in dedicated hardware are necessary to fully realize this potential.
\end{abstract}


\section{Introduction}
Deep neural networks (DNNs) are fundamental to numerous recent advancements in Artificial Intelligence (AI), playing a central role in the emergence of foundation models and generative AI. One of the largest neural networks currently in use is the Megatron-Turing NGL 530B, a generative language model developed by Nvidia and Microsoft, boasting 530 billion parameters. However, deploying advanced machine learning models poses numerous complex engineering challenges due to the need for powerful computational devices and large memory storage. This not only affects training but also renders it unfeasible to run on devices with limited computational resources \cite{Dai_2023}.

The pioneering works by \cite{Gupta2015DeepLW} and \cite{Courbariaux_2015} originated from the idea that leveraging the inherent noise-tolerance of neural network algorithms could allow for the relaxation of certain constraints on underlying hardware. \cite{Gupta2015DeepLW} specifically explored the use of low-precision fixed-point arithmetic for training deep neural networks, with a particular focus on the rounding mode used during operations on fixed-point numbers. \cite{Courbariaux_2015} analysed the performance of a Maxout network in three benchmark datasets with three distinct formats: floating point, fixed point and dynamic fixed point. The authors compared the performance with results from literature.

Lower precision means each number uses fewer bits, which makes memory smaller, cheaper, and more power-efficient. Fixed-point arithmetic is simpler and faster than floating-point, especially with low precision. This reduces the load on processors, allowing them to perform more operations per second and improving overall efficiency. Using less power extends battery life in portable devices and lowers cooling needs, reducing system costs and complexity. Lower precision can lead to more affordable and accessible hardware, enabling wider use of neural network applications.

Inspired on the works by \cite{Gupta2015DeepLW} and \cite{Courbariaux_2015}, it is natural to quantize parameters to optimize memory usage and improve computational efficiency, instead of reducing their total number in large AI models. Quantization decreases the bit width of the values used, thereby reducing the computational cost of floating-point operations. For example, in \cite{Moosmann_2024}, the authors explored the accuracy and suitability for real-time applications of a fully quantized ultra-lightweight object detection network.  

Although quantization can significantly decrease computational cost and memory usage, the information loss can potentially lower overall model performance, particularly in terms of accuracy. The trade-off between quantization and model performance is a balance between achieving greater efficiency and maintaining acceptable levels of accuracy. Most algorithms used for quantizing parameters of AI models require a pre-trained model with at least 32-bit precision. Once a pre-trained model is obtained, techniques such as {\em Post-Training Quantization} (\cite{Liu_2021,gong_2024}) and {\em Quantization Aware Training} (\cite{KIRTAS2022561,Lima_2022,siddegowda_2022}) are applied to ensure effective quantization. However, the lower the quantization, meaning the fewer bits used, the more degraded the model's performance becomes. Consequently, these methods typically do not use fewer than 8 bits to represent parameters after quantization. This limitation results in minimal reduction of the memory required to store the model and provides little significant improvement in computational efficiency.  In \cite{YANG202070}, a methodology to train ResNet models with full 8-bit integers is presented. 

In \cite{CHU2021107647} the authors address the trade-off between model performance and computational efficiency by adopting a mixed-precision approach. Their proposal is based on the fact that different layers within neural networks contribute differently to overall performance and vary in their sensitivity to quantization.

Binarization is a 1-bit quantization where data can only have $-1$ or $+1$ values, and the ideia of {\em Binarized Neural Networks}  open the possibility of a new era of more efficient neural network models that require less memory and can be applied to various types of problem \cite{QIN2020107281}. The works by \cite{Courbariaux_2016} and \cite{Hubara_2016} introduced an efficient method for training BNNs involving binary weights and activations. During the forward pass, weights and activations are binarized to either $-1$ or $+1$, reducing memory usage and enabling faster bitwise operations to replace most arithmetic computations. However, binary functions lack gradients necessary for backpropagation during training. To address this, real-valued weights are utilized to compute gradients and update parameters, facilitated by a hard hyperbolic tangent function that provides continuous and differentiable gradients. This approach ensures effective propagation of gradients through the network during backpropagation, enhancing training efficiency. Real-valued weights are subsequently binarized for use in the forward pass, maintaining computational efficiency during forward and inference while providing accurate gradient computation and parameter updates in the training phase. Furthermore, the authors devised a binary matrix multiplication GPU kernel to accelerate execution of the binary network compared to using an unoptimized GPU kernel. However, the reported findings are confined to a single model with convolutional layers and do not address scalability or extension of their proposals.

At the same time, \cite{Rastegari_2016} studied two different approaches on large-scale datasets like ImageNet: the traditional BNNs with all weight values are approximated using binary values; and XNOR-Networks, that extends the first concept by also approximating inputs with binary values. The authors claimed that the last approach provided $\approx 58\times$ speed up and enabled the possibility of running the inference of state of
the art deep neural network on CPU in real-time.

In 2018, \cite{DENG201849} introduced a novel approach for training DNNs. They addressed memory and computation bottlenecks by proposing a method to backpropagate through discrete activations and eliminate full-precision hidden weights during training. The authors constrained weights and activations in the ternary space $-1,0,+1$ to form what they called gated XNOR networks.

In \cite{Wang_2023}, the authors introduced a novel approach for large language models, training the model from scratch with quantization, diverging from the common practice of applying quantization post-training. They developed BitNet, which is a 1-bit transformer architecture designed for large language models (llmS). BitNet employs low-precision binary weights and quantized activations, while maintaining high precision for the optimizer states and gradients during training. The results from \cite{Wang_2023} demonstrate that BitNet achieves competitive performance while substantially reducing memory usage compared to 8-bit quantization methods and 16-bit precision transformers.

In \cite{IGNATOV2020276,PENG201991} and \cite{Tang_2020}, different network binarization approaches  were proposed to solve the lower prediction accuracy by using binary weights and fast bitwise operations. More recently, \cite{Ma_2024} introduced a low-bit variant of llm, named BitNet 1.58, where each model parameter is ternary ${-1, 0, +1}$. This model achieves equivalent performance to models with the same number of parameters but trained with 16-bit precision, using the same number of tokens. Thus, BitNet 1.58 offers significant advantages including lower latency, reduced memory footprint, and decreased energy consumption.  In \cite{Ma_2024}, the authors suggested that the 1.58-bit LLM model sets a new benchmark and paradigm for training next-generation high-performance and cost-effective LLMs. 

The previous contributions create opportunities for more efficient neural network models that require less memory and can be applied to various types of problems. Additionally, these advancements support the development of specialized hardware optimized for low-bit neural networks.

This study study examines the bit precision required for model parameters to match the performance of 32-bit models in multiclass image classification. It evaluates quantization levels from 1 to 4.08 bits, analyzing their impact on neural network accuracy and efficiency. Lower-bit quantization (1-1.5 bits) reduces memory and computation but may harm accuracy, while higher-bit levels (3.17-4.08 bits) balance efficiency and performance. The study considers fully connected layers (FCNN), convolutional layers (CVNN), and transformer blocks (Visual Transformer - VIT) models, identifying quantization strategies that optimize computational resources for deployment in edge computing and embedded systems. Section \ref{sec:Quantization} outlines the parameter quantization method employed, Section \ref{sec:DataTrainingParam} presents the data and training parameters, and Sections \ref{sec:FCNN}, \ref{sec:CVNN}, and \ref{sec:VIT} respectively detail the FCNN, CVNN, and VIT models, along with comparative results across different parameter resolutions. Section 7 introduces a method for storing weights using fewer bits of resolution. Finally, Section 8 summarizes the conclusions drawn from this study.


\section{Quantization method}
\label{sec:Quantization}

Since the model weights are the parameters that demand the most memory and computational resources in neural networks—both in convolutional layers and fully connected and attention layers—only the weights of the connections are constrained to use low-resolution parameters in the models. The biases of all layers retain a 32-bit resolution.

During training, the connection weights are stored with 32-bit resolution, but the layer activations are calculated using weights at the specified bit resolution. Therefore, a quantization method is required for the weights during the training process. The desired number of discrete values is first selected, for example:

\begin{description}
    \item[1-bit resolution] the model weights are constrained to $2$ values: $-1;+1$
    \item[1.5-bit resolution] the model weights are constrained to $3$ values: $-1;0;+1$;
    \item[2-bit resolution] the model weights are constrained to $4$ values: $-1;-0,3333;+0.3333;+1$;
    \item[2.32-bit resolution] the model weights are constrained to $5$ values: $-1;-0.5;0;+0.5;+1$;
\end{description}
and so on.

It is important to note that, to reduce the memory required for storing model weights, they can be represented as positive integers, with their bit resolution determined by the defined number of possible discrete values. Weight quantization to the desired resolution is performed layer by layer in the model. Equations (\ref{eq:QuantProcess_1}) to (\ref{eq:QuantProcess_4}) implement the parameter quantization.

\begin{equation}\label{eq:QuantProcess_1}
    v_{max} = \frac{N_{values}-1}{2}
\end{equation}
\begin{equation}\label{eq:QuantProcess_2}
    \mathbf W_{norm} = \frac{\mathbf W}{\beta \bar{W}}
\end{equation}

\begin{equation}\label{eq:QuantProcess_3}
    \mathbf W_q = \frac{\texttt{round}\left(\mathbf W_{norm}v_{max}+v_{max}\right)-v_{max}}{v_{max}}
\end{equation}

\begin{equation}\label{eq:QuantProcess_4}
      \mathbf W_q =\begin{cases}
               +1,\quad\text{ if } \mathbf W_q > +1-\frac{1}{N_{values}} \\
               -1,\quad\text{ if } \mathbf W_q < -1+\frac{1}{N_{values}} \\
            \end{cases}
\end{equation}
where $N_{values}$ is the number of desired values for the layer weights, $\mathbf W$ is the tensor of weights of the layer, $\bar{W}$ is the mean of the layer weights, $\mathbf W_{norm}$ is the tensor of weights normalized by the mean, $1 \leq \beta \leq 2$ is a parameter for regulating the distribution of quantized weight values, $\mathbf W_q$ is the tensor of quantized weights, and $\texttt{round}$ is a function performing rounding operation. The value of $\beta$ in this work is set to $1.4$ because this value ensures a uniform distribution among the three weight values when $N_{values}$ equals $3$.

As previously mentioned, during model training, weights are maintained as 32-bit real values. However, in the forward propagation calculations, quantized weights are employed, computed according to equations (\ref{eq:QuantProcess_1}) to (\ref{eq:QuantProcess_4}). This approach is essential because if weights were stored in their quantized form, during training the parameter updates would be eliminated by the quantization function, hindering learning. Note that, most updates during training are typically on the order of $10^{-4}$ to $10^{-2}$, which would round to zero in the quantization process, preventing the original parameters from being updated and thus hindering the model's learning. To address this issue, we employ a technique inspired by the implementation of Vector Quantised-Variational AutoEncoder (VQ-VAE) proposed by \cite{Oord_2017}. This method, which ensures effective parameter updates despite quantization, is detailed in Section \ref{sec:FCNN}.


\section{Data and training parameters} \label{sec:DataTrainingParam}

Analyzing the number of bits required for a model to achieve the performance of models with 32-bit resolution weights constitutes a comparative study where the dataset used does not significantly influence the conclusions. Therefore, the CIFAR-10 dataset \cite{Krizhevsky_2009} is employed. This dataset features a straightforward multiclass classification task with low-resolution color images ($32\times 32\times 3$ pixels) across $10$ object classes. It is split into a training set with $50,000$ images and a test set with $10,000$ images.

A critical aspect of AI models is their generalization capability. Training with original data often leads to overfitting. To evaluate the generalization capacity of models with low-resolution weights, data augmentation is also used during training. Minor transformations are applied to the images to avoid significant distortion, including the following:
\begin{itemize}
    \item horizontal shift: $\pm 10\%$;
    \item vertical shift: $\pm 10\%$;
    \item zoom in/out: $\pm 20\%$;
    \item horizontal flip (left/right);
    \item rotation: $\pm 5^o$.
\end{itemize}

In addition to using data augmentation to mitigate overfitting, models were trained with dropout layers. However, the inclusion of dropout was found to be ineffective in reducing overfitting, so results from models with dropout are not presented. The hyperparameters used in training the models are listed in Table \ref{tab:Hyperparam}.

\begin{table}[!t]
\caption{Hyperparameters used in model training.}
\label{tab:Hyperparam}
\centering
\begin{tabularx}{\linewidth}{XX}
\toprule
Hyperparameter & Value\\
\midrule
Cost function &	Categorical Cross Entropy\\
Optimization method &	Stochastic Gradient Descendent with Momentum\\
Momentum rate &	0.92\\
Metrics &	Accuracy\\
Learning rate for fully connected and convlutional models & 0.001 (fixed)\\
Learning rate for models with transformer blocks & 0.01 (fixed)\\
Batch size &	256\\
Number of epochs with original data for fully connected and convolutional models &	200\\
Number of epochs with data augmentation for fully connected and convolutional models &	1000\\
Number of epochs with original data for models with transformer blocks &	300\\
Number of epochs with data augmentation for models with transformer blocks &	2000\\
\bottomrule
\end{tabularx}
\end{table}


\section{Models with only fully connected layers}
\label{sec:FCNN}

Algorithm \ref{alg:foward_FCL} illustrates the forward propagation process in a fully connected layer with low-resolution connection weights. The function $\texttt{quant}$ performs the weight quantization defined by equations (\ref{eq:QuantProcess_1}) to (\ref{eq:QuantProcess_4}), $\texttt{no\_gradient}$ is a placeholder function that prevents gradient calculation for its argument during model training, $\texttt{activation}$ represents the chosen activation function for the layer, and $b$ is the bias vector of the layer.

\begin{algorithm*}
\caption{Forward propagation calculation process in a fully connected layer with low-resolution weights.}\label{alg:foward_FCL}
\begin{algorithmic}[1]
\REQUIRE Input $\mathbf x$, weights $\mathbf W$, bias $\mathbf b$, scaling factor $\beta$, weight mean $\bar W$, trainable flag $trainable$
\ENSURE Activations $\mathbf{a}$
\STATE Calculate weights adjust factor: $\gamma = \beta \bar{W}$
\STATE Calculate normalized weights: $\mathbf {W}_{norm} = \frac{\mathbf W}{\gamma}$
\IF{$trainable$}
    \STATE Quantize weights for training: $\mathbf {W}_q = \mathbf {W}_{norm} + \texttt{no\_gradient}(\texttt{quant}(\mathbf{W}_{norm}) - \mathbf{W}_{norm})$
\ELSE
    \STATE Quantize weights for inference: $\mathbf{W}_q = \texttt{quant}(\mathbf {W}_{norm})$
\ENDIF
\STATE Calculate activations: $\mathbf{a} = \texttt{activation}(\gamma \mathbf {W}_q \cdot \mathbf{x} + \mathbf b)$
\RETURN $\mathbf {a}$
\end{algorithmic}
\end{algorithm*}

The quantized weights used to compute layer activations remain consistent during both training and inference. However, during training, when $trainable$ variable is set to True, the calculated gradients and corresponding weight adjustments are stored in the normalized weight matrix ($W_{norm}$), ensuring no information is lost. This method allows for weight updates during training without sacrificing precision. The technique of adjusting parameters without information loss while using quantized weights in forward propagation calculations was adapted from \cite{Alcorn_2023}.

\subsection{Configuration of the models with fully connected layers}

Two simple models are configured with differences in the number of layers and units per layer. Algorithm 2 outlines the forward propagation process for the simpler fully connected layers model (FCNN1), while Algorithm 3 presents the more complex model with a larger number of parameters (FCNN2).

\begin{algorithm*}
\caption{Forward propagation process in the simpler fully connected layers model (FCNN1).} \label{alg:foward_FCNN1}
\begin{algorithmic}[1]
\REQUIRE Input image $\mathbf{x}$,  number of weight values $N_{values}$
\ENSURE Predicted output $\mathbf{ypred}$
\STATE $\mathbf {xf} = \texttt{Flatten}(\mathbf{x})$
\STATE $\mathbf{a1} = \text{FCL}(\text{units}=512, N_{values}, \text{activation}=\texttt{relu})(\mathbf{xf})$
\STATE $\mathbf{a2} = \text{FCL}(\text{units}=256, N_{values}, \text{activation}=\texttt{relu})(\mathbf{a1})$
\STATE $\mathbf{a3} = \text{FCL}(\text{units}=128, N_{values}, \text{activation}=\texttt{relu})(\mathbf{a2})$
\STATE $\mathbf{ypred} = \text{FCL}(\text{units}=10, N_{values}, \text{activation}=\texttt{softmax})(\mathbf{a3})$
\end{algorithmic}
\end{algorithm*}

\begin{algorithm*}
\caption{Forward propagation process in the more complex fully connected layers model (FCNN2).} \label{alg:foward_FCNN2}
\begin{algorithmic}[1]
\REQUIRE Input image $\mathbf{x}$,  number of weight values $N_{values}$
\ENSURE Predicted output $\mathbf{ypred}$
\STATE $\mathbf{xf} = \texttt{Flatten}(\mathbf{x})$
\STATE $\mathbf{a1} = \text{FCL}(\text{units}=1024, N_{values}, \text{activation}=\texttt{relu})(\mathbf{xf})$
\STATE $\mathbf{a2} = \text{FCL}(\text{units}=512, N_{values}, \text{activation}=\texttt{relu})(\mathbf{a1})$
\STATE $\mathbf{a3} = \text{FCL}(\text{units}=256, N_{values}, \text{activation}=\texttt{relu})(\mathbf{a2})$
\STATE $\mathbf{a4} = \text{FCL}(\text{units}=128, N_{values}, \text{activation}=\texttt{relu})(\mathbf{a3})$
\STATE $\mathbf{ypred} = \text{FCL}(\text{units}=10, N_{values}, \text{activation}=\texttt{softmax})(\mathbf{a4})$
\end{algorithmic}
\end{algorithm*}

In Algorithms \ref{alg:foward_FCNN1} and  \ref{alg:foward_FCNN2}, $\text{FCL}$ represents the fully connected layer defined in Algorithm \ref{alg:foward_FCL}, and $\texttt{Flatten}$ is a function that converts an image into a vector. The number of units in the output layers is set to $10$ due to the presence of $10$ object classes. Given the multiclass classification problem, the output layer employs a $\texttt{softmax}$ activation function, while $\texttt{relu}$ is used in other layers. The number of values used for each weight, which is proportional to the number of bits, is determined by the parameter $N_{values}$. In both models, weight connections and layer biases are initialized using standard methods: {\em Glorot Uniform } for weights and zeros for biases. No regularization techniques or parameter constraints are applied.

The number of units used in the layers of the simpler model (FCNN1) are respectively $512, 256, 128$, and $10$, totaling $1,738,890$ weights and biases. In the more complex model (FCNN2), the numbers of units in the layers are respectively $512, 256, 128$, and $10$, totaling $3,837,066$ weights and biases.

\subsection{The results of the models with fully connected layers}
\label{sec:ResultsFC}

We utilize models with low-resolution weights set to $2, 3, 4, 5, 8, 9, 16$, and $17$ different values, corresponding to resolutions of $1, 1.5, 2, 2.32, 3, 3.17, 4$, and $4.08$ bits per weight, respectively. It is important to clarify that the reported results are examples from individual training runs. However, extensive repetitions were conducted for each model under identical conditions, consistently producing similar outcomes. Therefore, presenting a single example result for each model is highly indicative.

\begin{figure*}[hbt!]
\begin{subfigure}{.475\linewidth}
  \includegraphics[width=\linewidth]{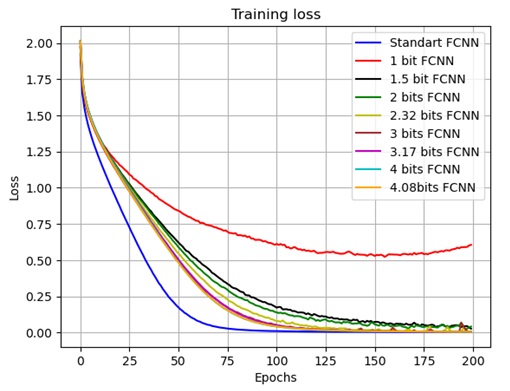}
  \caption{Training Loss}
  \label{FCNN1_TrainingLos}
\end{subfigure}\hfill 
\begin{subfigure}{.475\linewidth}
  \includegraphics[width=\linewidth]{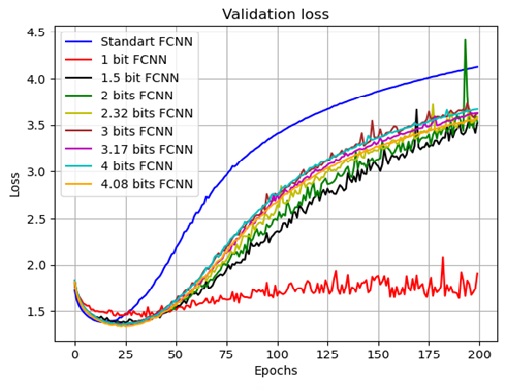}
  \caption{Validation Loss}
  \label{FCNN1_ValidationLoss}
\end{subfigure}
\medskip 
\begin{subfigure}{.475\linewidth}
  \includegraphics[width=\linewidth]{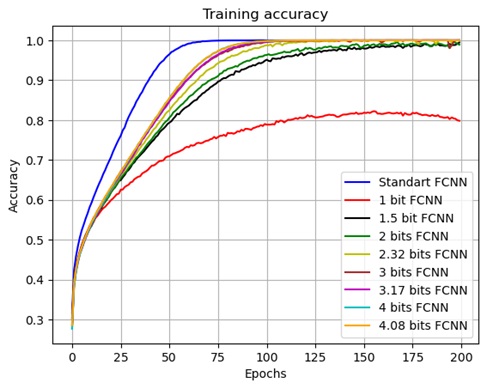}
  \caption{Training Accuracy}
  \label{FCNN1_TrainingAccuracy}
\end{subfigure}\hfill 
\begin{subfigure}{.495\linewidth}
  \includegraphics[width=\linewidth]{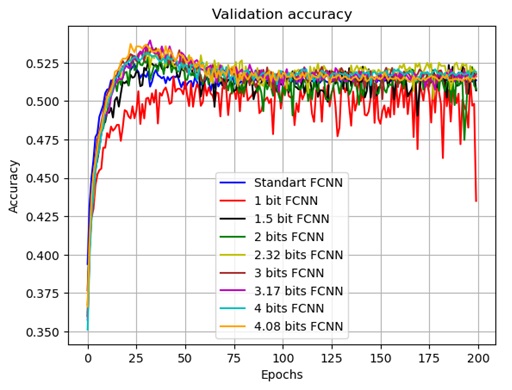}
  \caption{Validation Accuracy}
  \label{FCNN1_ValidationAccuracy}
\end{subfigure}
\caption{Training and validation results for the simpler models with only fully connected layers (FCNN1) for various resolutions used in the weights.}
\label{fig:FCNN1_TrainValid}
\end{figure*}

\begin{figure*}[hbt!]
\begin{subfigure}{.475\linewidth}
  \includegraphics[width=\linewidth]{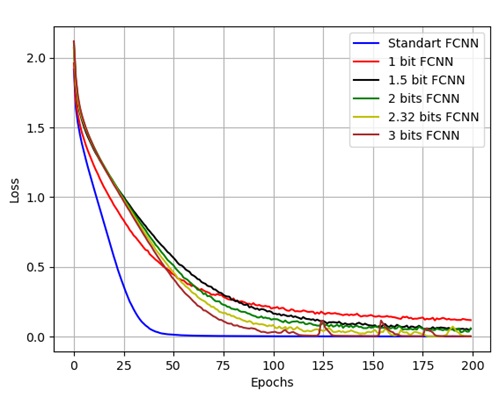}
  \caption{Training Loss}
  \label{FCNN2_TrainingLoss}
\end{subfigure}\hfill 
\begin{subfigure}{.475\linewidth}
  \includegraphics[width=\linewidth]{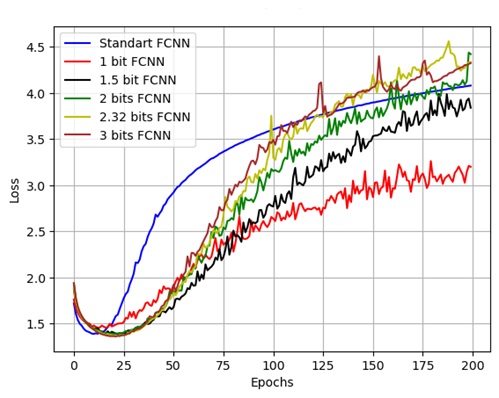}
  \caption{Validation Loss}
  \label{FCNN2_ValidationLoss}
\end{subfigure}
\medskip 
\begin{subfigure}{.475\linewidth}
  \includegraphics[width=\linewidth]{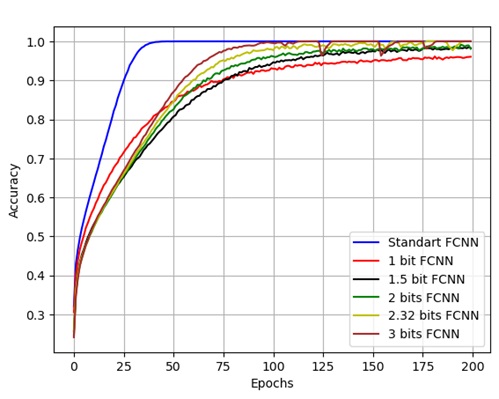}
  \caption{Training Accuracy}
  \label{FCNN2_TrainingAccuracy}
\end{subfigure}\hfill 
\begin{subfigure}{.475\linewidth}
  \includegraphics[width=\linewidth]{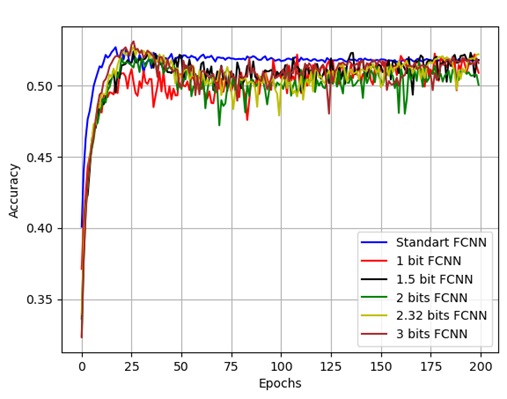}
  \caption{Validation Accuracy}
  \label{FCNN2_ValidationAccuracy}
\end{subfigure}
\caption{Training and validation results for the more complex models with only fully connected layers (FCNN2) for various resolutions used in the weights.}
\label{fig:FCNN2_TrainValid}
\end{figure*}

Figure \ref{fig:FCNN1_TrainValid} shows the training results for the models featuring only fully connected layers with the simpler configuration (FCNN1), while Figure \ref{fig:FCNN2_TrainValid} presents the results for the more complex models with various resolutions used for the weights. The results of a standard 32-bit parameter network is included as a benchmark reference.

The results depicted in Figures \ref{fig:FCNN1_TrainValid} and \ref{fig:FCNN2_TrainValid} reveal several crucial findings:

\begin{enumerate}
    \item Training the simpler model (FCNN1) with 1-bit weights ($N_{values} = 2$) proves ineffective, resulting in unsatisfactory performance.
    \item Models with lower-resolution parameters require more epochs to achieve adequate training.
    \item Models with a higher number of parameters in the low-resolution configuration progressively approach the performance level of the standard model.
    \item All models, including the standard model, demonstrate overfitting issues, leading to inferior performance on validation data compared to training data.
\end{enumerate}

\begin{figure*}[hbt!]
\begin{subfigure}{.475\linewidth}
  \includegraphics[width=\linewidth]{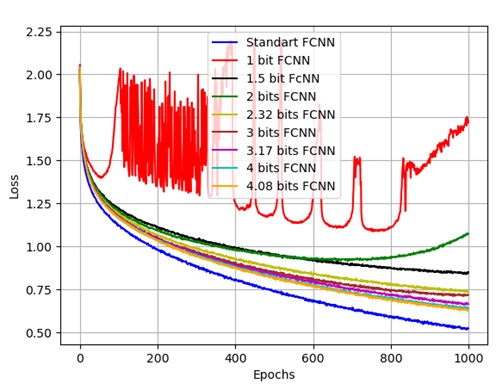}
  \caption{Training Loss}
  \label{FCNN1_AUG_TrainingLos}
\end{subfigure}\hfill 
\begin{subfigure}{.475\linewidth}
  \includegraphics[width=\linewidth]{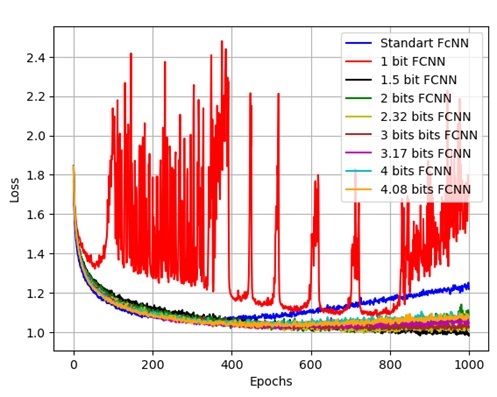}
  \caption{Validation Loss}
  \label{FCNN1_AUG_ValidationLoss}
\end{subfigure}
\medskip 
\begin{subfigure}{.475\linewidth}
  \includegraphics[width=\linewidth]{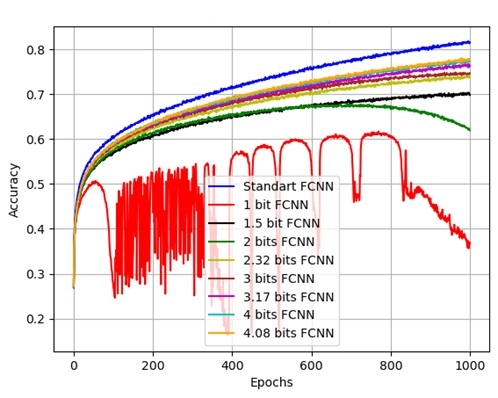}
  \caption{Training Accuracy}
  \label{FCNN1_AUG_TrainingAccuracy}
\end{subfigure}\hfill 
\begin{subfigure}{.495\linewidth}
  \includegraphics[width=\linewidth]{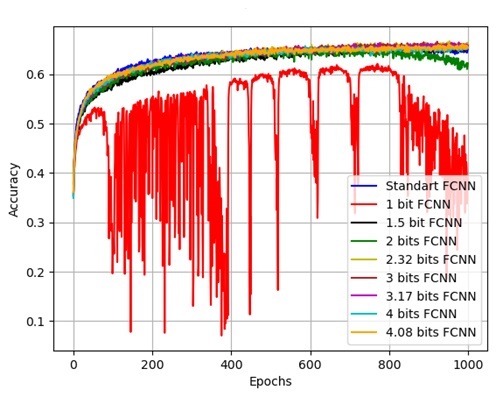}
  \caption{Validation Accuracy}
  \label{FCNN1_AUG_ValidationAccuracy}
\end{subfigure}
\caption{Training and validation results for the simpler models with only fully connected layers (FCNN1) using data augmentation for various resolutions used in the weights.}
\label{fig:FCNN1_AUG_TrainValid}
\end{figure*}

\begin{figure*}[hbt!]
\begin{subfigure}{.475\linewidth}
  \includegraphics[width=\linewidth]{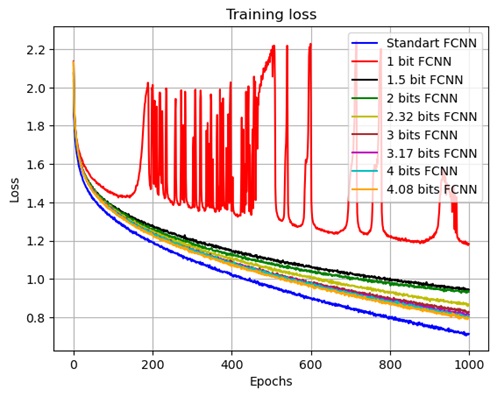}
  \caption{Training Loss}
  \label{FCNN2_AUG_TrainingLoss}
\end{subfigure}\hfill 
\begin{subfigure}{.475\linewidth}
  \includegraphics[width=\linewidth]{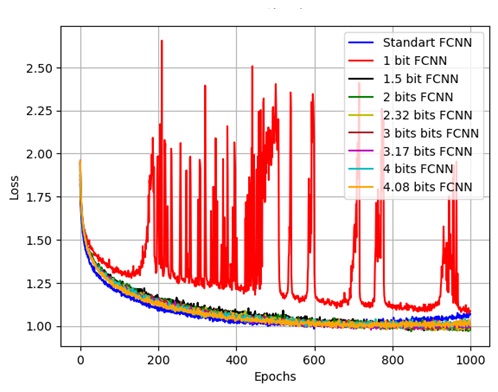}
  \caption{Validation Loss}
  \label{FCNN2_AUG_ValidationLoss}
\end{subfigure}
\medskip
\begin{subfigure}{.475\linewidth}
  \includegraphics[width=\linewidth]{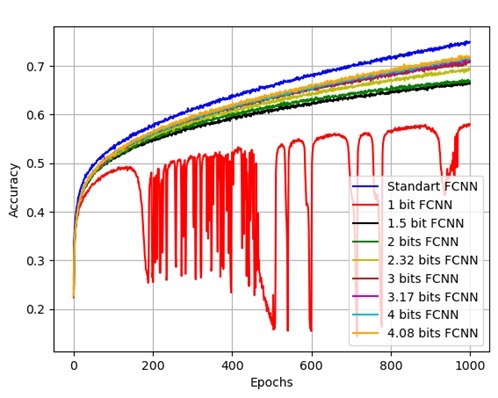}
  \caption{Training Accuracy}
  \label{FCNN2_AUG_TrainingAccuracy}
\end{subfigure}\hfill 
\begin{subfigure}{.475\linewidth}
  \includegraphics[width=\linewidth]{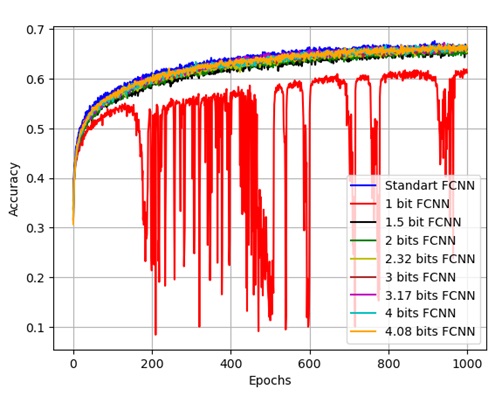}
  \caption{Validation Accuracy}
  \label{FCNN2_AUG_ValidationAccuracy}
\end{subfigure}

\caption{Training and validation results for the more complex models with only fully connected layers (FCNN2) using data augmentation for various resolutions used in the weights.}
\label{fig:FCNN2_AUG_TrainValid}
\end{figure*}

Figures \ref{fig:FCNN1_AUG_TrainValid} and \ref{fig:FCNN2_AUG_TrainValid} illustrate the training outcomes for both simpler (FCNN1) and more complex (FCNN2) models with data augmentation over 1000 training epochs. The specific image transformations used during training are detailed in Section \ref{sec:DataTrainingParam}.

It is important to note that multiple training runs were conducted for all models, consistently yielding similar results.

Upon analyzing the training results of the fully connected neural network models with data augmentation, as depicted in Figures \ref{fig:FCNN1_AUG_TrainValid} and \ref{fig:FCNN2_AUG_TrainValid}, several insights emerge:
\begin{enumerate}
    \item Both models utilizing 1-bit weights exhibit training instability and produce unsatisfactory results.
    \item Interestingly, the 1.5-bit model performs better than the 2-bit model despite the latter having higher resolution, suggesting that the inclusion of zero among possible weight values plays a crucial role.
    \item The more complex model with 2-bit weights achieves satisfactory results and performs comparably to models with higher-resolution weights, indicating that a large number of parameters allows effective learning even with lower-resolution weights.
    \item Similar to training without data augmentation, models with lower-resolution weights require a greater number of epochs to achieve results comparable to those of standard 32-bit models.
\end{enumerate}

Note that all models that do not exhibit training instability show overfitting, albeit to a lesser degree than observed in training without data augmentation. Therefore, this shows that models with low resolution weights are capable of generalization in the same way as 32-bit models.

\section{Models with convolutional layers}
\label{sec:CVNN}

The only difference between the convolutional layer and the fully connected layer with low-resolution weights is that in computing the activations, convolution operation is used between the filters (connection weights) and the input tensor of the layer, rather than matrix multiplication. Equation \ref{eq:Activ_CVNN} performs convolution operation in calculating activations for convolutional layers with low resolution weights. 

\begin{equation}\label{eq:Activ_CVNN}
   \mathbf a = \texttt{activation}\left(\texttt{conv}\left(\gamma \mathbf W_q,\mathbf x\right)+\mathbf b\right)    
\end{equation}
where $\texttt{conv}\left(\gamma \mathbf W_q,\mathbf x\right)$ performs two dimensional convolution of $\mathbf x$ by $\gamma \mathbf W_q$. It should be noted that Equation (\ref{eq:Activ_CVNN}) replaces the activation calculation in Algorithm \ref{alg:foward_FCL}, which implements the forward propagation process in fully connected layers.

\subsection{Configuration of the models with convolutional layers}
\label{sec:ConfigCNN}

Two models are configured with differences in the number of layers and units per layer. Algorithm 4 outlines the forward propagation calculation for the simpler convolutional layer model (CVNN1), while Algorithm 5 details the calculation process for the more complex model (CVNN2).

\begin{algorithm*}
\caption{Forward propagation process for the simpler convolutional layer model (CVNN1).} \label{alg:foward_CCNN1}
\begin{algorithmic}[1]
\REQUIRE Input image $\mathbf{x}$, number of weight Values $N_{values}$
\ENSURE Predicted output $\mathbf{ypred}$
\STATE $\mathbf{a1} = \text{Conv2D}(\text{units}=64, N_{values}, \text{activation}=\texttt{relu}, \text{padding}=\texttt{same})(\mathbf{x})$
\STATE $\mathbf{a1} = \text{MaxPool2D}((2,2), \text{strides}=(2,2))(\mathbf{a1})$
\STATE $\mathbf{a2} = \text{Conv2D}(\text{units}=128, N_{values}, \text{activation}=\texttt{relu}, \text{padding}=\texttt{same})(\mathbf{a1})$
\STATE $\mathbf{a2} = \text{MaxPool2D}((2,2), \text{strides}=(2,2))(\mathbf{a2})$
\STATE $\mathbf{a3} = \text{Conv2D}(\text{units}=256, N_{values}, \text{activation}=\texttt{relu}, \text{padding}=\texttt{same})(\mathbf{a2})$
\STATE $\mathbf{a3} = \text{MaxPool2D}((2,2), \text{strides}=(2,2))(\mathbf{a3})$
\STATE $\mathbf{a4} = \text{Flatten}(\mathbf{a3})$
\STATE $\mathbf{a5} = \text{FCL}(\text{units}=128, N_{values}, \text{activation}=\texttt{relu})(\mathbf{a4})$
\STATE $\mathbf{ypred} = \text{FCL}(\text{units}=10, N_{values}, \text{activation}=\texttt{softmax})(\mathbf{a5})$
\end{algorithmic}

\end{algorithm*}

\begin{algorithm*}
\caption{Forward propagation process for the more complex convolutional layer model (CVNN2).} \label{alg:foward_CCNN2}
\begin{algorithmic}[1]
\REQUIRE Input image $\mathbf{x}$, number of weight values $\mathbf{N_{values}}$
\ENSURE Predicted output $\mathbf{ypred}$
\STATE $\mathbf{a1} = \text{Conv2D}(\text{units}=128, N_{values}, \text{activation}=\texttt{relu}, \text{padding}=\texttt{same})(\mathbf{x})$
\STATE $\mathbf{a1} = \text{Conv2D}(\text{units}=128, N_{values}, \text{activation}=\texttt{relu}, \text{padding}=\texttt{same})(\mathbf{a1})$
\STATE $\mathbf{a1} = \text{MaxPool2D}((2,2), \text{strides}=(2,2))(\mathbf{a1})$
\STATE $\mathbf{a2} = \text{Conv2D}(\text{units}=256, N_{values}, \text{activation}=\texttt{relu}, \text{padding}=\texttt{same})(\mathbf{a1})$
\STATE $\mathbf{a2} = \text{Conv2D}(\text{units}=256, N_{values}, \text{activation}=\texttt{relu}, \text{padding}=\texttt{same})(\mathbf{a2})$
\STATE $\mathbf{a2} = \text{MaxPool2D}((2,2), \text{strides}=(2,2))(\mathbf{a2})$
\STATE $\mathbf{a3} = \text{Conv2D}(\text{units}=512, N_{values}, \text{activation}=\texttt{relu}, \text{padding}=\texttt{same})(\mathbf{a2})$
\STATE $\mathbf{a3} = \text{Conv2D}(\text{units}=512, N_{values}, \text{activation}=\texttt{relu}, \text{padding}=\texttt{same})(\mathbf{a3})$
\STATE $\mathbf{a3} = \text{MaxPool2D}((2,2), \text{strides}=(2,2))(\mathbf{a3})$
\STATE $\mathbf{a4} = \text{Flatten}(\mathbf{a3})$
\STATE $\mathbf{a5} = \text{FCL}(\text{units}=512, N_{values}, \text{activation}=\texttt{relu})(\mathbf{a4})$
\STATE $\mathbf{ypred} = \text{FCL}(\text{units}=10, N_{values}, \text{activation}=\texttt{softmax})(\mathbf{a5})$
\end{algorithmic}
\end{algorithm*}

In Algorithms \ref{alg:foward_CCNN1} and \ref{alg:foward_CCNN2}, Conv2D denotes a two-dimensional convolutional layer, while MaxPool2D signifies a two-dimensional max-pooling layer. Each convolutional layer employs 3x3 filters with a $\texttt{relu}$ activation function, a stride of 1, and padding to preserve the input tensor dimensions in the output tensors. The first fully connected layer also uses $\texttt{relu}$ activation, whereas the output layer adopts $\texttt{softmax}$ activation. The parameter $N_{values}$ defines the number of possible values for the connection weights.

In both models, connection weights and biases are initialized using standard methods: {\em Glorot Uniform } for weights and zeros for biases. No regularization techniques or parameter constraints are applied. The simpler model (CVNN1) comprises 896,522 parameters, while the more complex model (CVNN2) includes 8,776,330 parameters, accounting for both weights and biases.

\subsection{Results obtained with the models with convolutional layers}

\begin{figure*}[hbt!]
\begin{subfigure}{.475\linewidth}
  \includegraphics[width=\linewidth]{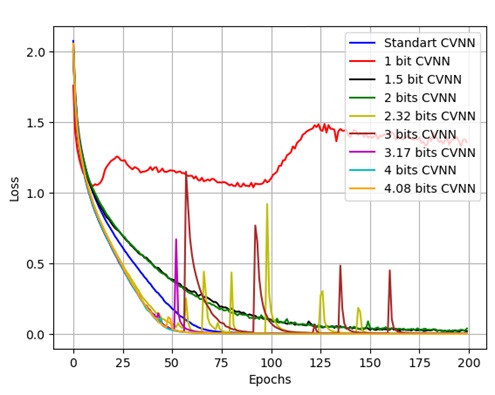}
  \caption{Training Loss}
  \label{CVNN1_TrainingLos}
\end{subfigure}\hfill 
\begin{subfigure}{.475\linewidth}
  \includegraphics[width=\linewidth]{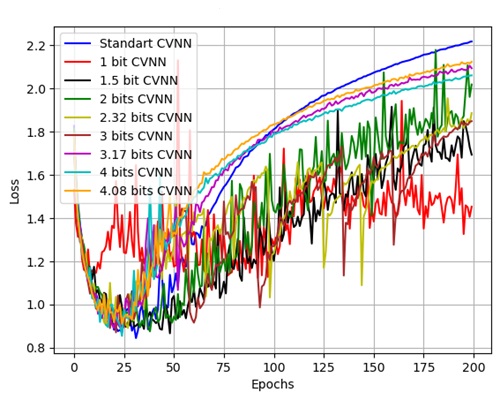}
  \caption{Validation Loss}
  \label{CVNN1_ValidationLoss}
\end{subfigure}
\medskip 
\begin{subfigure}{.475\linewidth}
  \includegraphics[width=\linewidth]{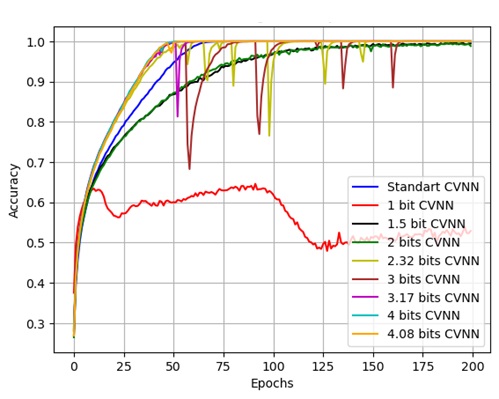}
  \caption{Training Accuracy}
  \label{CVNN1_TrainingAccuracy}
\end{subfigure}\hfill 
\begin{subfigure}{.495\linewidth}
  \includegraphics[width=\linewidth]{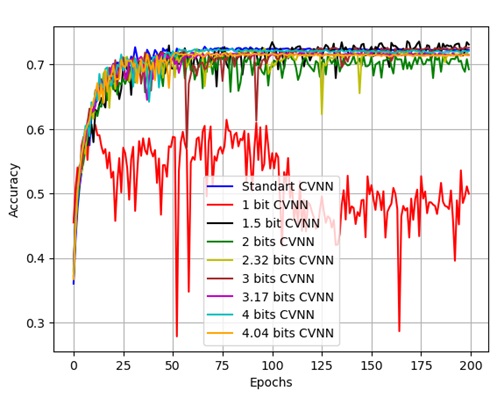}
  \caption{Validation Accuracy}
  \label{CVNN1_ValidationAccuracy}
\end{subfigure}
\caption{Training and validation results for the simpler models with Convolutional layer models (CVNN1) for various resolutions used in the weights.}
\label{fig:CVNN1_TrainValid}
\end{figure*}

\begin{figure*}[hbt!]
\begin{subfigure}{.475\linewidth}
  \includegraphics[width=\linewidth]{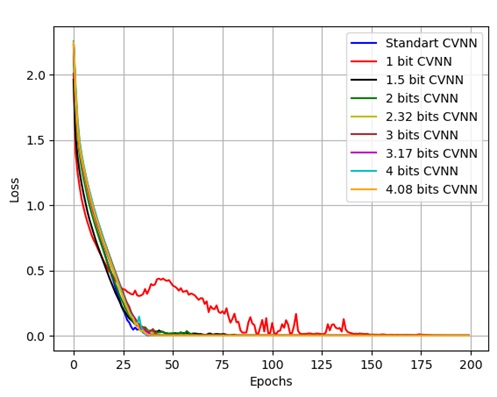}
  \caption{Training Loss}
  \label{CVNN2_TrainingLoss}
\end{subfigure}\hfill 
\begin{subfigure}{.475\linewidth}
  \includegraphics[width=\linewidth]{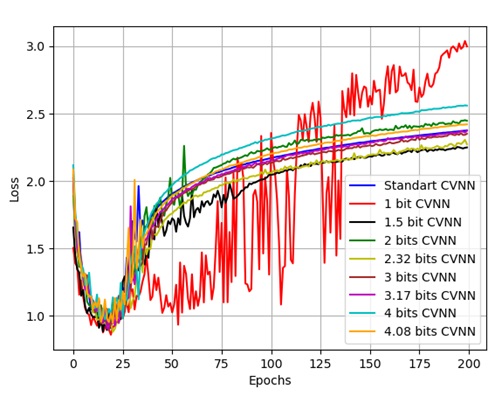}
  \caption{Validation Loss}
  \label{CVNN2_ValidationLoss}
\end{subfigure}
\medskip 
\begin{subfigure}{.475\linewidth}
  \includegraphics[width=\linewidth]{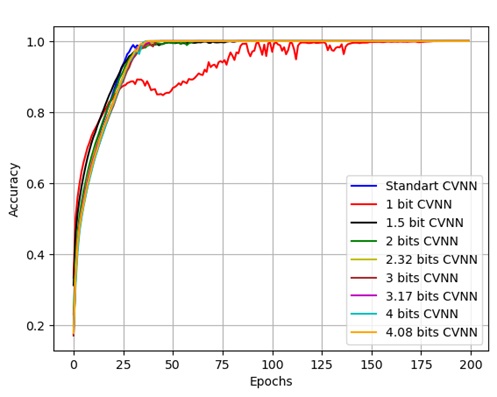}
  \caption{Training Accuracy}
  \label{CVNN2_TrainingAccuracy}
\end{subfigure}\hfill 
\begin{subfigure}{.475\linewidth}
  \includegraphics[width=\linewidth]{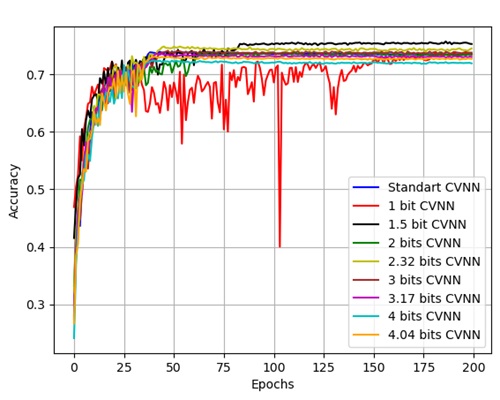}
  \caption{Validation Accuracy}
  \label{CVNN2_ValidationAccuracy}
\end{subfigure}
\caption{Training and validation results for the more complex models with convolutional layer models (CVNN2) for various resolutions used in the weights.}
\label{fig:CNN2_TrainValid}
\end{figure*}

In Figure \ref{fig:CVNN1_TrainValid}, the training results are displayed for the simpler convolutional layer models (CVNN1), while Figure \ref{fig:CNN2_TrainValid} presents the results for the more complex models (CVNN2), both for various resolutions used in the layer weights. Once again, results from standard networks with 32-bit parameters are included as a benchmark for the desired performance. It is important to note that multiple training tests were conducted for all models, and all results are very similar.

Analyzing the training results of the convolutional models shown in Figures 5 and 6, one can observe that the results of the low-resolution models at the end of training, except for the simpler 1-bit model (CVNN1), are nearly identical to those of the standard models. Other observations can be done:
\begin{itemize}
    \item The simpler model (CVNN1) with 1-bit weights exhibits instability throughout all training processes and fails to learn the data, unlike the more complex 1-bit model, which despite some instability during training, is able to learn effectively;
    \item The simpler models (CVNN1) with 1.5 and 2-bit resolutions require more epochs to achieve comparable results compared to the standard 32-bit model. Conversely, models with resolutions of 2.32, 3, 3.17, 4, and 4.08 bits among the simpler CVNN1 models require fewer training epochs than the standard model. Also, the low-resolution more complex models (CVNN2) with more than 1-bit require the same number of epochs to train as the standard model;
    \item The simpler low-resolution models (CVNN1) exhibit occasional sharp fluctuations in the cost function and accuracy, which quickly stabilize. These fluctuations are likely associated with simultaneous changes in a large number of weights. Notably, these oscillations do not occur during training of the more complex models (CVNN2). 
\end{itemize}

\begin{figure*}[hbt!]
\begin{subfigure}{.475\linewidth}
  \includegraphics[width=\linewidth]{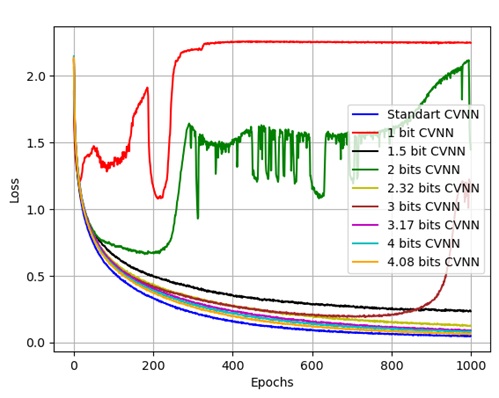}
  \caption{Training Loss}
  \label{CVNN1_AUG_TrainingLos}
\end{subfigure}\hfill 
\begin{subfigure}{.475\linewidth}
  \includegraphics[width=\linewidth]{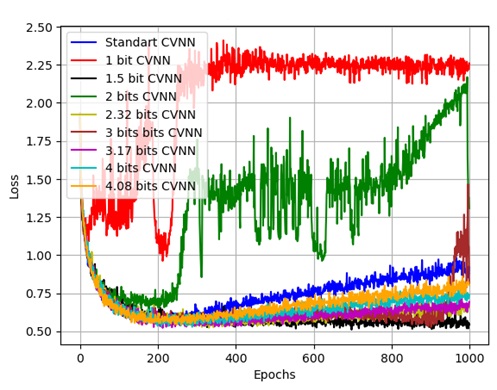}
  \caption{Validation Loss}
  \label{CVNN1_AUG_ValidationLoss}
\end{subfigure}
\medskip 
\begin{subfigure}{.475\linewidth}
  \includegraphics[width=\linewidth]{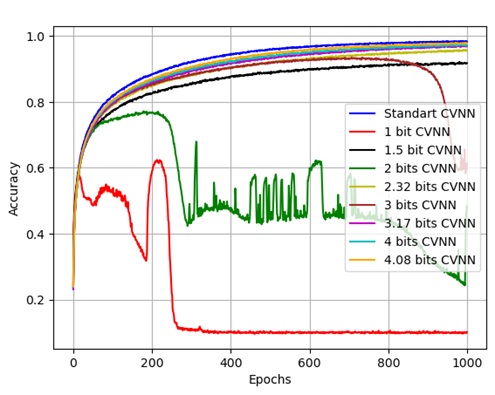}
  \caption{Training Accuracy}
  \label{CVNN1_AUG_TrainingAccuracy}
\end{subfigure}\hfill 
\begin{subfigure}{.495\linewidth}
  \includegraphics[width=\linewidth]{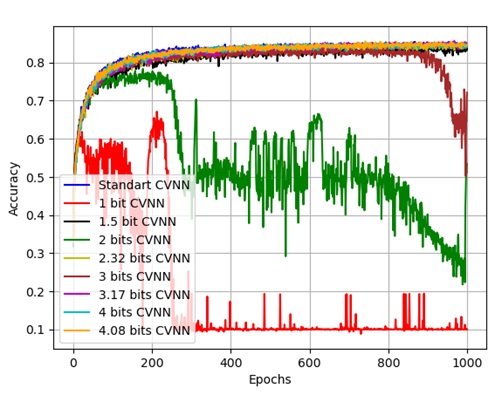}
  \caption{Validation Accuracy}
  \label{CVNN1_AUG_ValidationAccuracy}
\end{subfigure}
\caption{Training and validation results for the simpler models with only fully connected layers (FCNN1) using data augmentation for various resolutions used in the weights.}
\label{fig:CVNN1_AUG_TrainValid}
\end{figure*}

\begin{figure*}[hbt!]
\begin{subfigure}{.475\linewidth}
  \includegraphics[width=\linewidth]{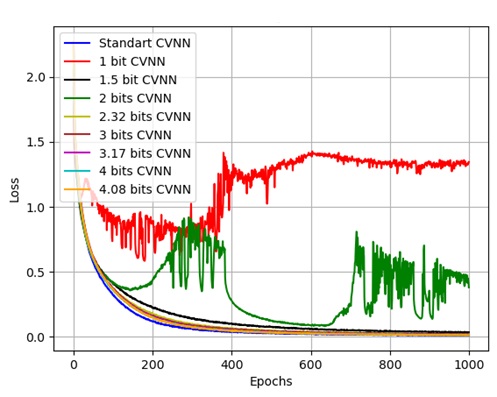}
  \caption{Training Loss}
  \label{CVNN2_AUG_TrainingLoss}
\end{subfigure}\hfill 
\begin{subfigure}{.475\linewidth}
  \includegraphics[width=\linewidth]{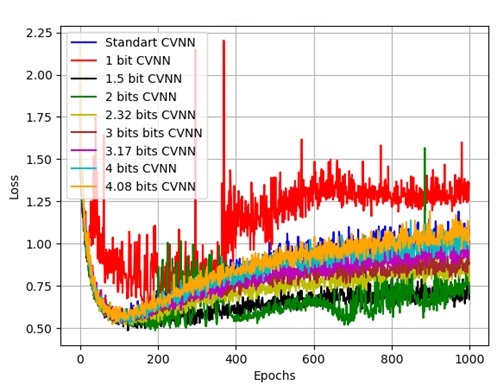}
  \caption{Validation Loss}
  \label{CVNN2_AUG_ValidationLoss}
\end{subfigure}
\medskip
\begin{subfigure}{.475\linewidth}
  \includegraphics[width=\linewidth]{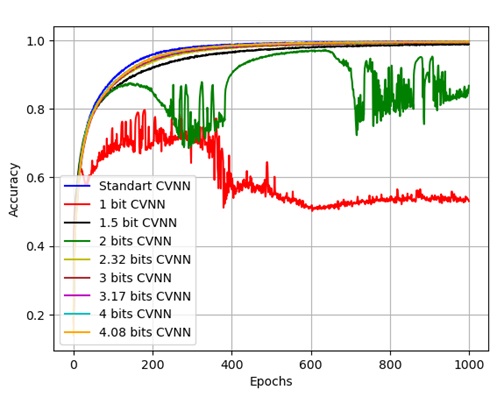}
  \caption{Training Accuracy}
  \label{CVNN2_AUG_TrainingAccuracy}
\end{subfigure}\hfill 
\begin{subfigure}{.475\linewidth}
  \includegraphics[width=\linewidth]{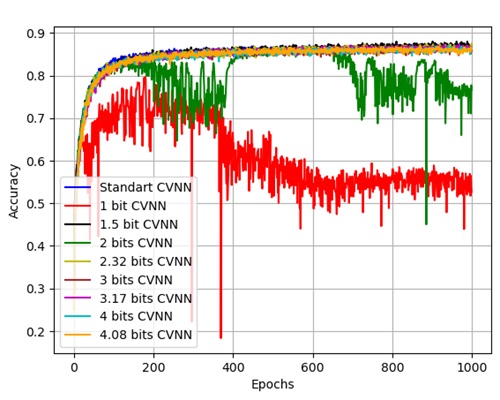}
  \caption{Validation Accuracy}
  \label{CVNN2_AUG_ValidationAccuracy}
\end{subfigure}
\caption{Training and validation results for the MORE complex convolutional layer models (CVNN2) using data augmentation for various resolutions used in the weights.}
\label{fig:CVNN2_AUG_TrainValid}
\end{figure*}

Excluding the simpler 1-bit model (CVNN1), all other models demonstrate overfitting problems. To verify the generalization ability of low-resolution convolutional models, training is repeated with data augmentation. The image transformations applied are identical to those used in the training of the models with only fully connected layers. Figure \ref{fig:CVNN1_AUG_TrainValid} presents the training results for the simpler models (CVNN1), while Figure \ref{fig:CVNN2_AUG_TrainValid} displays the results for the more complex models (CVNN2), using 1000 training epochs. Multiple training runs were conducted for all models, and consistent results were observed across all experiments.

Analyzing the results presented in Figures 7 and 8 reveals that data augmentation tends to induce instability in the training of low-resolution weight models. All the low resolutions models with 1, 2 and 3 bits exhibit training instability and unsatisfactory results, thus none of these models trained with data augmentation yield satisfactory outcomes. Furthermore, except for the models experiencing instability, the simpler low-resolution models (CVNN1) generally show slightly worse results compared to the standard model, whereas the low-resolution more complex models (CVNN2) exhibit results similar to the standard model. An important observation is that models incorporating zero as a possible weight value, specifically the 1.5-bit, 2.32-bit, 3.17-bit, and 4.08-bit models, outperform models where zero is not included among the possible weight values.

To allow a better analysis of low-resolution convolutional layers, models with 5x5 filters are trained to investigate whether using larger filters impacts the performance of the model. It is important to note that these models are identical to the previously analyzed convolutional models, with the only modification being the filter size. Figure 9 shows the training results with data augmentation for the more complex models using 5x5 filters, employing 1000 training epochs. Multiple training sessions were conducted for all models, yielding consistent results across all the experiments.

\begin{figure*}[hbt!]
\begin{subfigure}{.475\linewidth}
  \includegraphics[width=\linewidth]{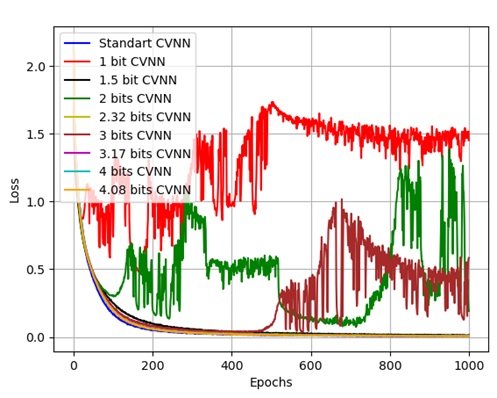}
  \caption{Training Loss}
  \label{CVNN2_AUG2_TrainingLoss}
\end{subfigure}\hfill 
\begin{subfigure}{.475\linewidth}
  \includegraphics[width=\linewidth]{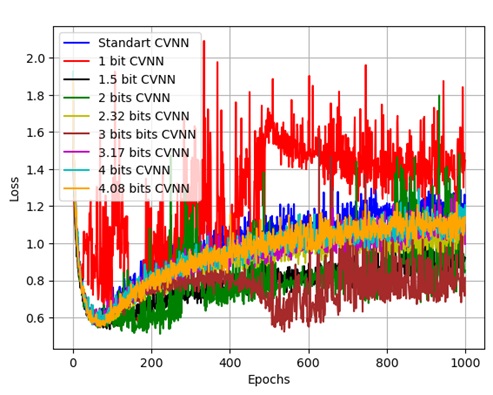}
  \caption{Validation Loss}
  \label{CVNN2_AUG2_ValidationLoss}
\end{subfigure}
\medskip
\begin{subfigure}{.475\linewidth}
  \includegraphics[width=\linewidth]{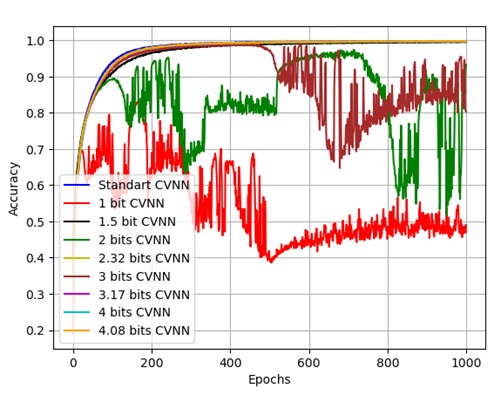}
  \caption{Training Accuracy}
  \label{CVNN2_AUG2_TrainingAccuracy}
\end{subfigure}\hfill 
\begin{subfigure}{.475\linewidth}
  \includegraphics[width=\linewidth]{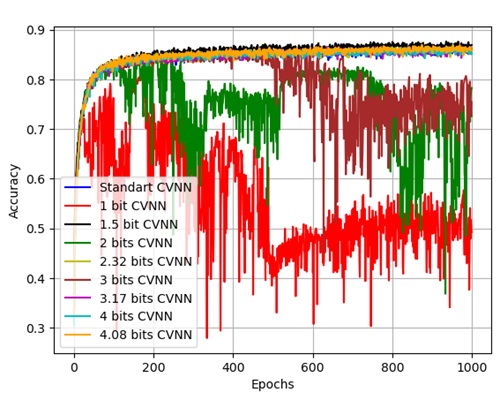}
  \caption{Validation Accuracy}
  \label{CVNN2_AUG2_ValidationAccuracy}
\end{subfigure}
\caption{Training and validation results for the MORE complex convolutional layer models (CVNN2) using data augmentation for various resolutions used in the weights.}
\label{fig:CVNN2_AUG2_TrainValid}
\end{figure*}

The results presented in Figure 9 shows that using 5x5 filters increases the issue of instability. In this case, aside from the more complex models (CVNN2) with 1-bit and 2-bit resolutions, the 3-bit model also exhibits instability. This result reinforces the observation that models with an even number of possible weight values tend to encounter more training difficulties compared to models with an odd number of possible weight values. Note that models with an odd number of possible values include weights that can be zero, whereas models with even numbers of possible values do not. This finding underscores the importance of including weights with values of zero, particularly for low-resolution weights and models with a few number of parameters.

\section{Models with transformer blocks (Visual Transformer - VIT)}
\label{sec:VIT} 

A transformer block, including its attention mechanism, primarily consists of embedding layers, fully connected layers, and normalization layers.  The fully connected layers of a transformer block with low-resolution weights are the same as those presented in Algorithm 1. Since the embedding and normalization layers have a small number of parameters compared to the fully connected layers, 32-bits parameters are used in these layers.

\subsection{Configuration of the visual transformer models}
\label{sec:ConfigVIT}

Two models that differ only in the number of transformer blocks and the units in each fully connected layer are used in this comparative study. Algorithm \ref{alg:foward_VIT} presents the forward propagation process in the VIT models.

\begin{algorithm*}
\caption{Forward propagation process of the VIT models.} \label{alg:foward_VIT}
\tiny
\begin{algorithmic}[1]
\REQUIRE Input image $\mathbf{x}$, patch dimension $patch\_size$, number of patches $num\_patches$ embedding dimension $emb\_dim$, number of transformer blocks $transformer\_layers$, vector with numbers of units in the fully connected layers of the transformer blocks $\mathbf{transformer\_units}$, number of weight values $N_{values}$, number of heads $num\_heads$
\ENSURE Predicted output $\mathbf{ypred}$
\STATE \# Create patches
\STATE $\mathbf{patches} = \texttt{Patches}(patch\_size)(\mathbf{x})$
\STATE \# Encode patches
\STATE $\mathbf{encoded\_patches} = \texttt{PatchEncoder}(num\_patches, emb\_dim)(\mathbf{patches})$
\STATE \# Create multiple layers of Transformer Blocks.
\FOR {$i = 1$ to $transformer\_layers$}
    \STATE \# Normalization layer 1
    \STATE $\mathbf{x1} = \texttt{LayerNormalization}(\mathbf{encoded\_patches})$
    \STATE \# Create a multi-head attention layer.
    \STATE $\mathbf{attention\_output}, \_ = \texttt{Attention}(emb\_dim, num\_heads, dropout\_rate=0.1)([\mathbf{x1}, \mathbf{x1}])$
    \STATE \# Skip connection 1
    \STATE $\mathbf{x2} = \texttt{Add}(\mathbf{attention\_output}, \mathbf{encoded\_patches})$
    \STATE \# Normalization layer 2
    \STATE $\mathbf{x3} = \texttt{LayerNormalization}(\mathbf{x2})$
    \STATE \# MLP
    \STATE $\mathbf{x3} = \text{FCL}(units=\mathbf{transformer\_units}[0], N\_values, activation=\texttt{gelu})(\mathbf{x3})$
    \STATE $matbf{x3} = \texttt{Dropout}(dropout\_rate=0.1)(\mathbf{x3})$
    \STATE $\mathbf{x3} = \text{FCL}(units=\mathbf{transformer\_units}[1], N\_values, activation=\texttt{gelu})(\mathbf{x3})$
    \STATE $\mathbf{x3} = \texttt{Dropout}(dropout\_rate=0.1)(\mathbf{x3})$
    \STATE \# Skip connection 2.
    \STATE $\mathbf{encoded\_patches} = \texttt{Add}(\mathbf{x3}, \mathbf{x2})$
\ENDFOR
\STATE \# Create a [batch\_size, projection\_dim] tensor.
\STATE $\mathbf{representation} = \texttt{LayerNormalization}(\mathbf{encoded\_patches})$
\STATE $\mathbf{representation} = \texttt{Flatten}(\mathbf{representation})$
\STATE $\mathbf{representation} = \texttt{Dropout}(dropout\_rate=0.5)(\mathbf{representation})$
\STATE \# MLP
\STATE $\mathbf{features} = \text{FCL}(units=\mathbf{mlp\_head\_units}[0], N\_values, activation=\texttt{gelu})(\mathbf{representation})$
\STATE $\mathbf{features} = \texttt{Dropout}(dropout\_rate=0.5)(\mathbf{features})$
\STATE $\mathbf{features} = \text{FCL}(units=\mathbf{mlp\_head\_units}[1], N\_values, activation=\texttt{gelu})(\mathbf{features})$
\STATE $\mathbf{features} = \texttt{Dropout}(dropout\_rate=0.5)(\mathbf{features})$
\STATE \# Classify outputs.
\STATE $\mathbf{ypred} = \text{FCL}(units=num\_classes, N\_values, activation=\texttt{softmax})(\mathbf{features})$
\end{algorithmic}
\end{algorithm*}

The components of Algorithm \ref{alg:foward_VIT} are detailed as follows:

\begin{itemize}
    \item \textbf{FLC} (Fully Connected Layer): This layer uses low-resolution weights, and its calculation process is described in Algorithm 1. As previously discussed, the number of possible values for the connection weights is determined by the parameter $N_{values}$. Additionally, the Gaussian Error Linear Unit (GELU) activation function is employed in certain layers to enhance non-linearity.

    \item \texttt{Patches}($patch\_size$): This function generates image patches with dimensions \((batch\_size, num\_patches\_h \times \text num\_patches\_w, patch\_size \times patch\_size \times channels)\). Here, $batch\_size$ represents the number of examples in a batch, $num\_patches\_h$ is the number of patches along the image height, $num\_patches\_w$ is the number of patches along the image width, $patch\_size$ is the size of each patch in pixels, and $channels$ indicates the number of image channels.

    \item \texttt{PatchEncoder}($num\_patches$, $emb\_dim$): This is a standard function used in visual transformers that performs the embedding encoding of image patches along with their respective positional encodings. The $num\_patches$ parameter denotes the total number of patches per image, while $emb\_dim$ represents the dimensionality of the embedding encoding for the patches. The encoding of image patches is carried out using a fully connected layer with a linear activation function and 32-bit weight resolution. Positional encoding is performed using a standard embedding layer.

    \item \texttt{LayerNormalization}: This is a standard layer that normalizes the features of each example, ensuring that the output remains stable and centered.

    \item \texttt{Dropout}(\texttt{dropout\_rate}): This is a standard dropout layer where \texttt{dropout\_rate} specifies the fraction of the input units to drop during training to prevent overfitting.

    \item \texttt{Attention}(\texttt{emb\_dim}, \texttt{num\_heads}, \texttt{dropout\_rate}): This component represents the standard attention mechanism utilized in visual transformer models. The arguments include \texttt{emb\_dim}, \texttt{dropout\_rate}, and \texttt{num\_heads}. The last argument indicates the number of attention heads. The calculation process for this attention mechanism is detailed in Algorithm \ref{alg:attention}. Within this algorithm, the \texttt{reshape} function resizes a tensor according to the specified dimensions, while the \texttt{permute} function rearranges the axes of a tensor based on the provided order. All other terms have been defined in earlier sections.
\end{itemize}

\begin{algorithm*}
\caption{Calculation process for Attention Mechanism}
\label{alg:attention}
\begin{algorithmic}[1]
\REQUIRE Inputs $\mathbf{x}$, embedding dimension $emb\_dim$, number of weight values $N_{values}$, number of heads $num\_heads$, dropout rate $dropout\_rate$
\ENSURE Predict output $\mathbf{output}$
\STATE \# Retrives query, key and value from input list
\STATE $\mathbf{query}, \mathbf{key}, \mathbf{value} \gets \mathbf{x}$
\STATE \# Calculate n\_key and batch\_size values
\STATE $n\_key = \frac{emb\_dim}{num\_heads}$
\STATE $batch\_size = \mathbf{key}.\texttt{shape}[0]$
\STATE \# Calculate Q, K and V matrices
\STATE $\mathbf{Q} = \text{FLC}(\text{units} = emb\_dim, N_{values}, \text{activation} = \texttt{linear}, \text{use\_bias} = \text{False})(\mathbf{query})$
\STATE $\mathbf{K} = \text{FLC}(\text{units} = emb\_dim, N_{values}, \text{activation} = \texttt{linear}, \text{use\_bias} = \text{False})(\mathbf{key})$
\STATE $\mathbf{V} = \text{FLC}(\text{units} = emb\_dim, N_{values}, \text{activation} = \texttt{linear}, \text{use\_bias} = \text{False})(\mathbf{value})$
\STATE \# Reshape and permute Q, K and V
\STATE $\mathbf{Q} = \texttt{reshape}(\mathbf{Q}, [batch\_size, -1, num\_heads, n\_key])$
\STATE $\mathbf{Q} = \texttt{permute}(\mathbf{Q}, [0, 2, 1, 3])$
\STATE $\mathbf{K} = \texttt{reshape}(\mathbf{K}, [batch\_size, -1, num\_heads, n\_key])$
\STATE $\mathbf{K} = \texttt{permute}(\mathbf{K}, [0, 2, 1, 3])$
\STATE $\mathbf{V} = \texttt{reshape}(\mathbf{V}, [batch\_size, -1, num\_heads, n\_key])$
\STATE $\mathbf{V} = \texttt{permute}(\mathbf{V}, [0, 2, 1, 3])$
\STATE \# Calculate dot product Q by K
\STATE $\mathbf{QK} = \texttt{matmul}(\mathbf{Q}, \texttt{permute}(\mathbf{K}, [0, 1, 3, 2])) / \sqrt{n\_key}$
\STATE \# Calculate attention probabilities 
\STATE $\mathbf{attn\_prob} = \texttt{softmax}(\mathbf{QK}, \text{axis} = -1)$
\STATE \# Calculate attention
\STATE $\mathbf{A} = \texttt{matmul}(\texttt{dropout}(dropout\_rate)(\mathbf{attn\_prob}), \mathbf{V})$
\STATE $\mathbf{A} = \texttt{permute}(\mathbf{A}, [0, 2, 1, 3])$
\STATE $\mathbf{A} = \texttt{reshape}(\mathbf{A}, [batch\_size, -1, num\_heads \cdot n\_key])$
\STATE \# Calculate output using FCL
\STATE $\mathbf{output} = \text{FLC}(\text{units} = emb\_dim, N_{values}, \text{activation} = \texttt{linear}, \text{use\_bias} = \text{False})(\mathbf{A})$
\end{algorithmic}
\end{algorithm*}

In Table \ref{tab:HyperparamVIT}, the hyperparameters used in the two VIT models implemented in this study are presented. For both models, the weights of the connections and biases of the fully connected layers are initialized using the standard method, i.e., {\em Glorot Uniform} initialization for weights and zeros for the biases where applicable. No regularization techniques are used, and there are no parameter constraints. The total number of parameters for the simpler model (VIT1) is 4,766,282, while for the more complex model (VIT2) it is 10,573,770.

\begin{table}[!t]
\caption{Hyperparameters used in model training}
\label{tab:HyperparamVIT}
\centering
\begin{tabularx}{\linewidth}{Xl}
\toprule
\textbf{Hyperparameter} & \textbf{Value} \\
\midrule
$patch\_size$ & $4$ \\
$num\_patches$  \\
$(\text{image\_size} / \text{patch\_size})^2$ & $64$ \\
$emb\_dim$ & $64$ \\
$num\_heads$ & $4$ \\
$transformer\_units$ \\
$[emb\_dim \times 2, emb\_dim]$ & $[128, 64]$ \\
$\mathbf{transformer\_layers}$ & $2$ (simpler model) \\
& $4$ (complex model) \\
$\mathbf{mlp\_head\_units}$ & $[1024, 512]$ (simpler model) \\
& $[2048, 1024]$ (complex model) \\
\text{Loss function} & \text{Categorical Cross Entropy} \\
\text{Optimization method} & \text{Gradient Descent with Momentum} \\
\bottomrule
\end{tabularx}
\end{table}

\subsection{Results obtained with the VIT models}
\label{sec:Results_VIT}

\begin{figure*}[hbt!]
\begin{subfigure}{.475\linewidth}
  \includegraphics[width=\linewidth]{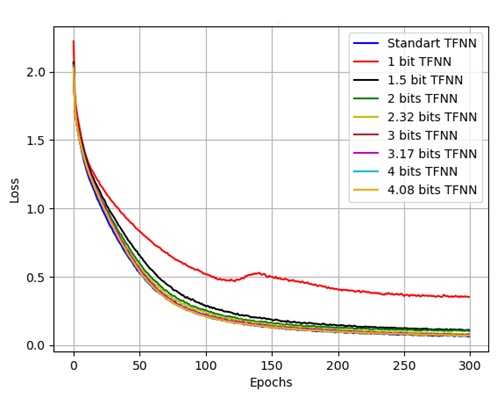}
  \caption{Training Loss}
  \label{VIT1_TrainingLoss}
\end{subfigure}\hfill 
\begin{subfigure}{.475\linewidth}
  \includegraphics[width=\linewidth]{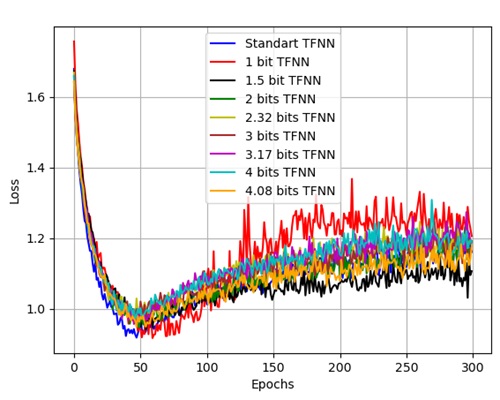}
  \caption{Validation Loss}
  \label{VIT1_ValidationLoss}
\end{subfigure}
\medskip
\begin{subfigure}{.475\linewidth}
  \includegraphics[width=\linewidth]{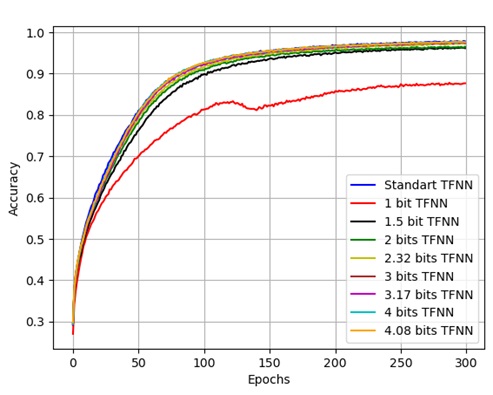}
  \caption{Training Accuracy}
  \label{VIT1_TrainingAccuracy}
\end{subfigure}\hfill 
\begin{subfigure}{.475\linewidth}
  \includegraphics[width=\linewidth]{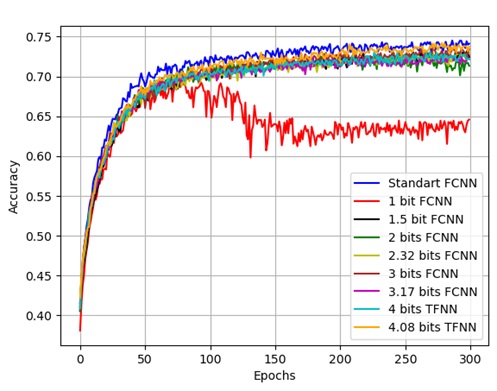}
  \caption{Validation Accuracy}
  \label{VIT1_ValidationAccuracy}
\end{subfigure}
\caption{Training and validation results for the simpler VIT models (VIT1) for various resolutions used in the weights.}
\label{fig:VIT1_TrainValid}
\end{figure*}

\begin{figure*}[hbt!]
\begin{subfigure}{.475\linewidth}
  \includegraphics[width=\linewidth]{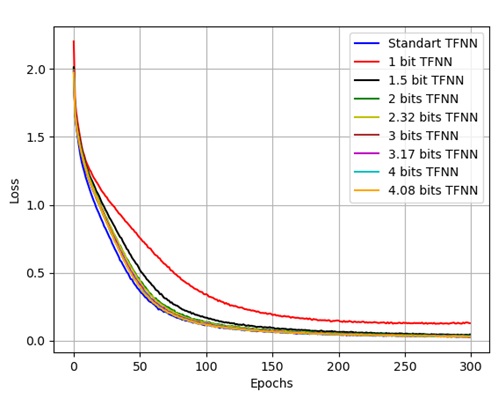}
  \caption{Training Loss}
  \label{VIT2_TrainingLoss}
\end{subfigure}\hfill 
\begin{subfigure}{.475\linewidth}
  \includegraphics[width=\linewidth]{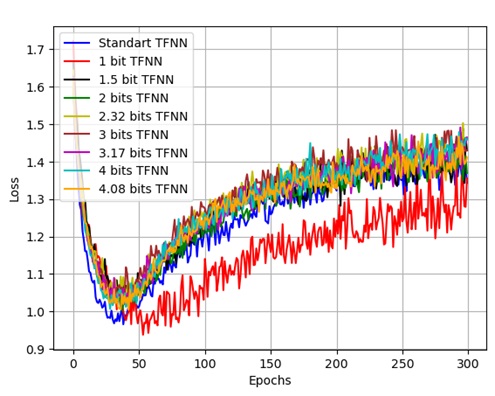}
  \caption{Validation Loss}
  \label{VIT2_ValidationLoss}
\end{subfigure}
\medskip
\begin{subfigure}{.475\linewidth}
  \includegraphics[width=\linewidth]{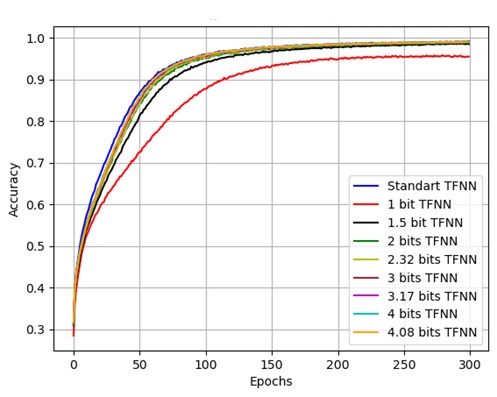}
  \caption{Training Accuracy}
  \label{VIT2_TrainingAccuracy}
\end{subfigure}\hfill 
\begin{subfigure}{.475\linewidth}
  \includegraphics[width=\linewidth]{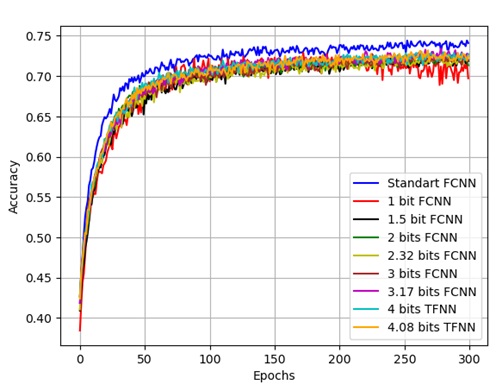}
  \caption{Validation Accuracy}
  \label{VIT2_ValidationAccuracy}
\end{subfigure}
\caption{Training and validation results for more complex VIT models (VIT2) for various resolutions used in the weights.}
\label{fig:VIT2_TrainValid}
\end{figure*}

Figure \ref{fig:VIT1_TrainValid} shows the training results for the simpler VIT models (VIT1) while Figure \ref{fig:VIT2_TrainValid} presents the results for the more complex VIT models (VIT2) for several resolutions used in the weights. Once again, the results for standard models with 32-bit parameters are provided as a reference for the desired performance. Again, it is important to note that multiple training tests were conducted for all models, and all results are highly consistent.

The training results of the VIT models presented in Figures \ref{fig:VIT1_TrainValid} and \ref{fig:VIT2_TrainValid}, show that the models with 1-bit weights exhibit poorer results compared to other models, but they are capable of learning the data, however, all other models, both simpler (VIT1) and more complex (VIT2), show results similar to those of the standard 32-bit resolution models. Also, for the VIT-type models, the parameter resolution appears to have less influence on the results quality compared to models with fully connected and convolutional layers.

\begin{figure*}[hbt!]
\begin{subfigure}{.475\linewidth}
  \includegraphics[width=\linewidth]{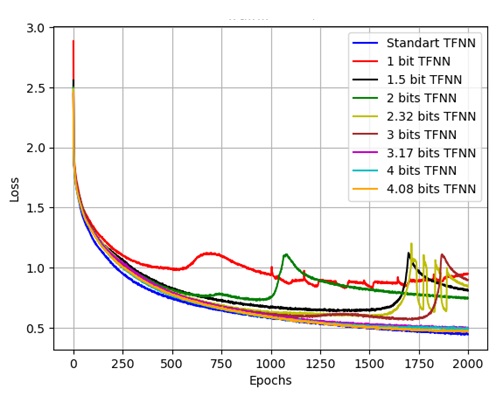}
  \caption{Training Loss}
  \label{VIT1_AUG_TrainingLoss}
\end{subfigure}\hfill 
\begin{subfigure}{.475\linewidth}
  \includegraphics[width=\linewidth]{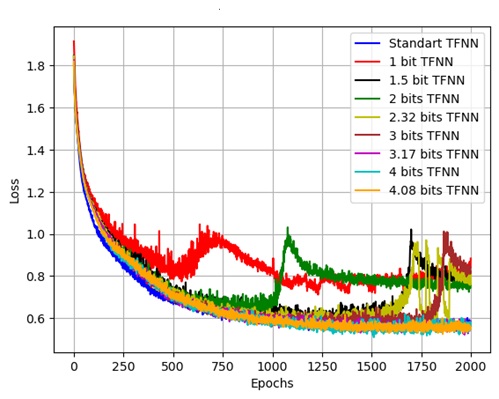}
  \caption{Validation Loss}
  \label{VIT1_AUG_ValidationLoss}
\end{subfigure}
\medskip
\begin{subfigure}{.475\linewidth}
  \includegraphics[width=\linewidth]{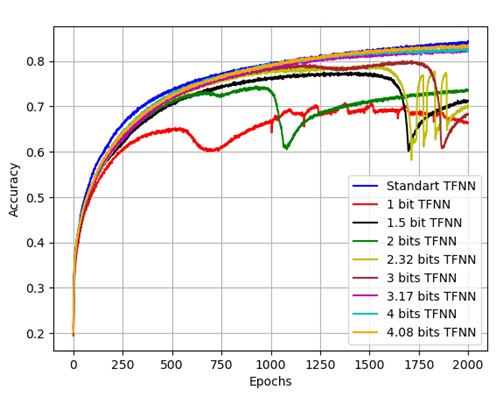}
  \caption{Training Accuracy}
  \label{VIT1_AUG_TrainingAccuracy}
\end{subfigure}\hfill 
\begin{subfigure}{.475\linewidth}
  \includegraphics[width=\linewidth]{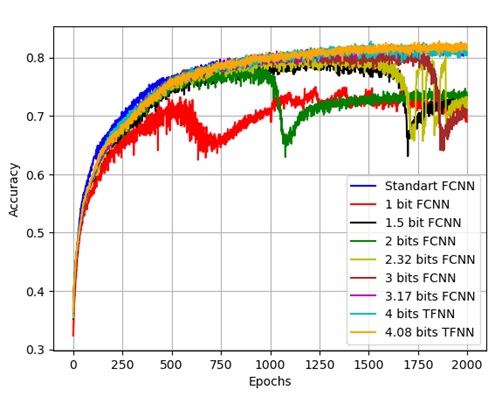}
  \caption{Validation Accuracy}
  \label{VIT1_AUG_ValidationAccuracy}
\end{subfigure}
\caption{Training and validation results for the simpler VIT models (VIT1) using data augmentation for various resolutions used in the weights.}
\label{fig:VIT1_AUG_TrainValid}
\end{figure*}

\begin{figure*}[hbt!]
\begin{subfigure}{.475\linewidth}
  \includegraphics[width=\linewidth]{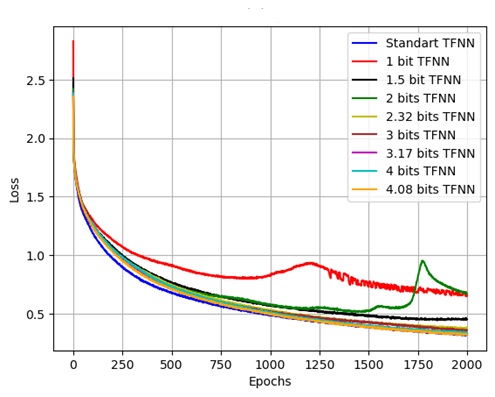}
  \caption{Training Loss}
  \label{VIT2_AUG_TrainingLoss}
\end{subfigure}\hfill 
\begin{subfigure}{.475\linewidth}
  \includegraphics[width=\linewidth]{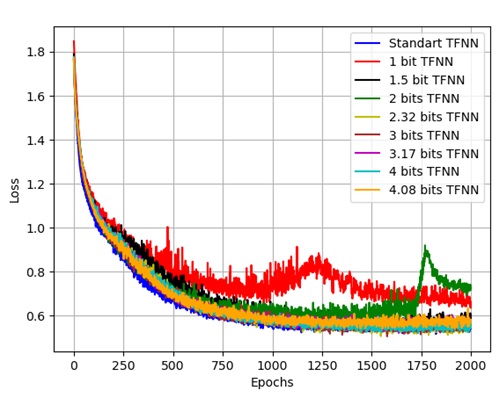}
  \caption{Validation Loss}
  \label{VIT2_AUG_ValidationLoss}
\end{subfigure}
\medskip
\begin{subfigure}{.475\linewidth}
  \includegraphics[width=\linewidth]{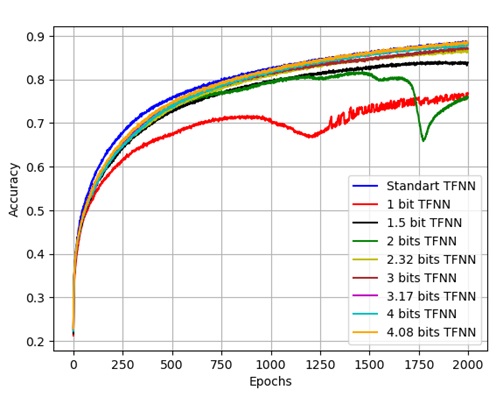}
  \caption{Training Accuracy}
  \label{VIT2_AUG_TrainingAccuracy}
\end{subfigure}\hfill 
\begin{subfigure}{.475\linewidth}
  \includegraphics[width=\linewidth]{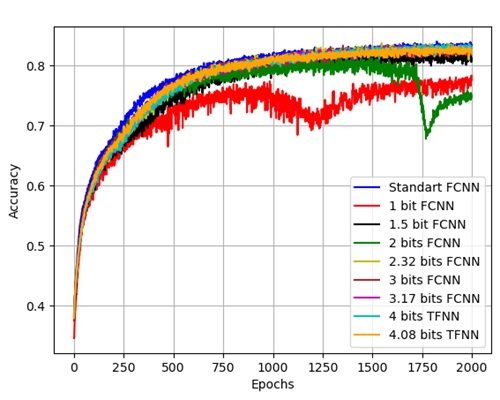}
  \caption{Validation Accuracy}
  \label{VIT2_AUG_ValidationAccuracy}
\end{subfigure}
\caption{Training and validation results for more complex VIT models (VIT2) using data augmentation for various resolutions used in the weights.}
\label{fig:VIT2_AUG_TrainValid}
\end{figure*}

All models exhibit overfitting problems regardless of the resolution of their weights. To verify the generalization capability of the low-resolution models, training is repeated with data augmentation. The image transformations applied are identical to those used in training models with fully connected and convolutional layers. Figure \ref{fig:VIT1_AUG_TrainValid} presents the training results for the simpler models (VIT1), while Figure \ref{fig:VIT2_AUG_TrainValid} presents the results for the more complex models (VIT2) with data augmentation across 2000 training epochs. It is noted that multiple trainings were conducted for all models and all results are consistent.

Results of Figures 12 and 13 show that training with data augmentation introduces instability in the results of several low-resolution models. For the simpler models (VIT1), instability occurs in models with 1, 1.5, 2, 2.32, and 3-bit resolution weights. Among the more complex models, instability is observed in models with 1 and 2-bit resolution weights. For the low-resolution models that do not exhibit training instability, the results are very similar to those obtained with standard 32-bit models. Furthermore, as expected, training with data augmentation reduces the problem of overfitting but does not completely eliminate it, however, the overfitting behavior of the low resolution models is the same as the standard 32-bit models.

Comparing the results of the VIT models with models using convolutional layers, it is observed that both types of models yield similar results. However, the VIT models appear to have less training instability without data augmentation and more instability with data augmentation. Additionally, the VIT models require a larger number of epochs to achieve comparable results.

\section{Memory Reduction for Models with Low-Resolution Weights}
\label{sec:MemoryReduction}

The architecture of current computers does not facilitate efficient multiplications involving both integers and real numbers. Therefore, for low-resolution weight models to achieve greater computational efficiency, it is essential to develop optimized hardware capable of performing operations with low-bit integers. Nevertheless, even with existing hardware, low-resolution models offer a significant advantage by requiring substantially less memory.

For instance, consider a model with 1.5-bit weights, which can take on three possible values: \(-1\), \(0\), and \(+1\). In this case, there are \(243\) (or \(3^5\)) possible combinations of five weights using these values. This implies that five weights of 1.5 bits can be stored within a single byte (8 bits). In comparison with storing weights in a 32-bit format (4 bytes), results in a memory reduction factor of 20. 

Table \ref{tab:MemoryReduction} presents the memory reduction for each of the low-resolution models analyzed in this study compared to 32-bit weights, assuming a byte is the minimum memory unit. It is important to note that this memory reduction only considers the weights of the connections.

\begin{table}[!t]
\caption{Reduction in memory usage of low-resolution weight models compared to 32-bit weight models}
\label{tab:MemoryReduction}
\centering
\begin{tabular}{cccc}
\toprule
\textbf{Number} & \textbf{Nvalues} & \textbf{Weights Stored} & \textbf{Memory} \\
\textbf{of Bits} &  & \textbf{by Byte} & \textbf{Reduction} \\
\midrule
1    & 2  & 1  & 32 \\
1.5  & 3  & 5  & 20 \\
2    & 4  & 4  & 16 \\
2.32 & 5  & 3  & 12 \\
3    & 8  & 2  & 8  \\
3.17 & 9  & 2  & 8  \\
4    & 16 & 2  & 8  \\
4.08 & 17 & 1  & 4  \\
\bottomrule
\end{tabular}
\end{table}

Analyzing the memory reduction achieved alongside the comparative performance of low-resolution weight models, we find that the optimal balance between performance and memory requirements is exhibited by the model with 2.32-bit weights (\(N_{\text{values}} = 5\)). This model demonstrates a 12-fold reduction in memory usage while maintaining stability during training across all model types, including fully connected layers, convolutional models, and transformer models.

\section{Conclusion}
In this study, we analyze the numbers of bits required for representing layer weights to achieve performance comparable to 32-bit resolution models. The focus is on multiclass object classification in images, examining models that utilize fully connected layers, convolutional layers, and transformer blocks, with weight resolutions ranging from 1 bit to 4.08 bits. 

Our approach deliberately avoids employing complex models aimed at maximizing performance or entirely eliminating overfitting. The primary goal is to determine whether low-resolution weight models can achieve similar performance levels to 32-bit models and generalize learning effectively through comparative analysis. It is important to note that no specialized algorithms were implemented to optimize the use of fewer bits in the weights, which remains an area for future exploration, especially with advancements in computer architectures capable of efficiently handling both small-bit integers and real numbers.

Several key conclusions emerge from this research:

\begin{enumerate}
    \item \textbf{Performance of Low-Resolution Models:} Initially, low-resolution models with small number of parameters yield results comparable to standard 32-bit models, although they require more training epochs. Despite this increased training time, low-resolution weight models can potentially be trained more rapidly, as calculations may be executed more efficiently due to the reduced bit representation. However, realizing this advantage requires the development of dedicated hardware capable of performing operations with numbers represented by fewer than 8 bits.

    \item \textbf{Impact of Data Augmentation:} Data augmentation appears to induce instability in the training of low-resolution weight models, particularly those with a small number of parameters. In contrast, models with a greater number of parameters demonstrate more stable training outcomes with data augmentation, especially those that accommodate zero as a possible weight value.

    \item \textbf{Preference for Odd Weight Values:} A significant finding is that using an odd number of possible weight values in low-resolution models—ensuring the inclusion of zero—yields better performance outcomes. Models with even \(N_{\text{values}}\), particularly those utilizing convolutional layers and transformer blocks, tend to exhibit training instability.

    \item \textbf{Computational Optimization:} Current computing systems are optimized for a minimum resolution of 8 bits (1 byte), indicating no computational speed advantage when comparing 1-bit to 8-bit resolution weights. However, training models with 8-bit weights can deliver performance comparable to models with 16 or 32-bit weights. It is crucial to note that post-training quantization from 16 or 32 bit weights to 8 bits often leads to performance degradation.

    \item \textbf{Advantages of Low-Resolution Models:} Low-resolution weight models present a significant advantage by enabling the development of more complex models with higher processing units while using substantially less memory compared to 32-bit resolution models or even quantized 8-bit models. These models hold great promise for facilitating the deployment of large language models in embedded devices.
\end{enumerate}

In summary, our findings indicate that using weights with 2.32 bits (\(N_{\text{values}} = 5\)) strikes the best balance between memory reduction, model performance, and efficiency. However, these conclusions should be regarded as preliminary and future studies should investigate other dataset types (language, time series, and image generation) and evaluate models with very large number of parameters (order of billions).


\bibliographystyle{unsrt}  
\bibliography{references}

\end{document}